\documentclass[journal]{IEEEtran}

\usepackage{amsmath,amssymb}
\usepackage{bm}
\usepackage{graphicx}
\usepackage{svg}
\usepackage{booktabs}
\usepackage{multirow}
\usepackage{array}
\usepackage{xcolor}
\usepackage{cite}
\usepackage{url}
\usepackage{microtype}
\usepackage{balance}
\usepackage{placeins}
\usepackage[hidelinks]{hyperref}
\hypersetup{
  breaklinks=true,
  bookmarksnumbered=true
}

\svgsetup{inkscapelatex=false}

\newcommand{\paperfigure}[4]{%
  \IfFileExists{#1.pdf}{%
    \includegraphics[width=#2]{#1.pdf}%
  }{\IfFileExists{#1.png}{%
    \includegraphics[width=#2]{#1.png}%
  }{\IfFileExists{#1.jpg}{%
    \includegraphics[width=#2]{#1.jpg}%
  }{\IfFileExists{#1.svg}{%
    \includesvg[width=#2]{#1}%
  }{%
    \fbox{\parbox[c][#3][c]{0.94\linewidth}{\centering\footnotesize #4}}%
  }}}}%
}

\begin{document}

\title{Reliability-Consistent Appearance Modeling for RGB--Infrared Multi-Object Tracking}

\author{Bingyang Wang,
Jiazuo Yu,
Lu Zhang,~\IEEEmembership{Member,~IEEE},
Lihe Zhang,
Huchuan Lu,~\IEEEmembership{Fellow,~IEEE},
and You He%
\thanks{This work has been submitted to the IEEE for possible publication. Copyright may be transferred without notice, after which this version may no longer be accessible.}%
\thanks{This work was supported in part by the National Natural Science Foundation of China under Grant Nos. 62441231 and 62293542; the Liao Ning Province Science and Technology Plan under Grant No. 2023JH26/10200016; the Dalian City Science and Technology Innovation Fund under Grant No. 2023JJ11CG001; and the Ningbo Key R\&D Project under Grant No. 2025Z039. (Corresponding author: Lu Zhang.)}%
\thanks{Bingyang Wang, Jiazuo Yu, Lu Zhang, Lihe Zhang, and Huchuan Lu are with the School of Information and Communication Engineering, Dalian University of Technology, Dalian 116024, China (e-mail: 22109095@mail.dlut.edu.cn; yujiazuo@mail.dlut.edu.cn; zhangluu@dlut.edu.cn; zhanglihe@dlut.edu.cn; lhchuan@dlut.edu.cn).}%
\thanks{You He is with the Department of Engineering, Tsinghua University, Beijing 100084, China (e-mail: youhe@mail.tsinghua.edu.cn).}%
}
\maketitle

\begin{abstract}
Recent RGB--infrared multi-object tracking (MOT) methods have substantially improved cross-modal representation learning and multimodal information integration. However, most existing methods compress RGB and infrared evidence into a unified descriptor. When one modality is temporarily degraded, the tracker can no longer regulate its contribution separately; unreliable evidence may distort association and contaminate appearance memory. We therefore propose RCA-MOT, a reliability-consistent appearance modeling framework that preserves cross-modal complementarity while retaining post-representation evidence controllability. RGB-specific, infrared-specific, and fused identity representations are maintained as jointly accessible appearance states. For each detection, a simple training-free target-scene estimator produces branch-wise appearance reliability used across the two temporal roles of appearance: relative reliability regulates different identity branches, absolute reliability calibrates borderline candidates near the ReID acceptance boundary, and the same branch-wise reliability controls how strongly matched descriptors update historical appearance memory. The individual components are intentionally simple, focusing evaluation on reliability-controllable appearance organization rather than elaborate fusion or learned reliability-estimation modules. RCA-MOT sets new SOTA results on VT-MOT and UniRTL, improving the previous best HOTA by 5.402/6.075 points and IDF1 by 11.549/12.789 points, respectively, while post-hoc analyses further validate reliability as an informative control signal. Code will be released upon acceptance.
\end{abstract}

\begin{IEEEkeywords}
RGB--infrared multi-object tracking, appearance reliability, data association,
appearance memory, multimodal tracking.
\end{IEEEkeywords}

\section{Introduction}
\label{sec:introduction}

\begin{figure}[!t]
\centering
\paperfigure{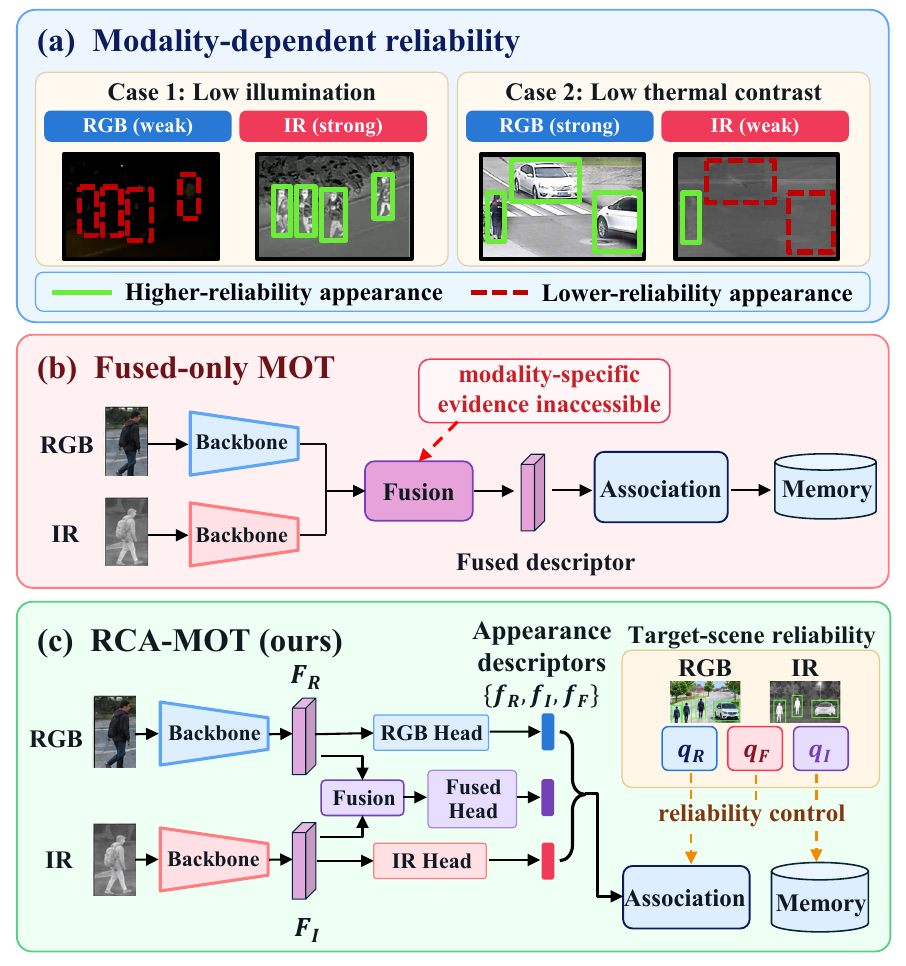}{0.8\columnwidth}{5.8cm}{Fig. 1 placeholder: modality-dependent reliability, fused-only limitation, and RCA-MOT conceptual comparison.}
\caption{Motivation for reliability-consistent appearance control. (a) Reliability is modality dependent: low illumination weakens RGB appearance, whereas low thermal contrast weakens infrared appearance. (b) Fused-only MOT makes modality-specific evidence inaccessible after fusion. (c) RCA-MOT retains RGB-specific, infrared-specific, and fused descriptors and uses target-scene reliability to regulate both association and appearance-memory update.}
\label{fig:motivation}
\end{figure}

RGB--infrared multi-object tracking (MOT) exploits complementary visible and thermal observations to preserve target identities under challenging illumination and visual degradation~\cite{pftrack,unirtl}. Existing RGB--infrared MOT methods have been developed mainly along a representation-centered route: cross-modal interaction and feature fusion are used to construct a more discriminative multimodal descriptor, which is then supplied to association and track maintenance. Representative systems such as PFTrack and UniS-MOT differ in their implementations, but broadly follow this fusion-centered paradigm. Consequently, recent work has substantially advanced how multimodal appearance is \emph{constructed}, yet has addressed less directly how the formed identity evidence should be used according to the condition of each observation. This omission matters because appearance in online MOT is not consumed only once; it participates in both the current identity decision and the evolving state of a trajectory.

This fusion-centered paradigm leaves an MOT-specific question unresolved: once a multimodal representation has been formed, how should its evidence be controlled when it is repeatedly used by the tracker? Fig.~\ref{fig:motivation}(a) illustrates why this matters at the observation level. Low illumination can selectively degrade RGB appearance while infrared remains informative, whereas low thermal contrast can produce the opposite reliability pattern. In a fused-only pipeline such as Fig.~\ref{fig:motivation}(b), however, modality-specific evidence becomes inaccessible once RGB and infrared information has been compressed into a single descriptor. The tracker may therefore retain a strong joint representation while losing the ability to regulate the underlying sources according to their current quality. This limitation is amplified by the recursive nature of MOT. Current appearance first affects the present track--detection association, and the descriptor of an accepted match may then enter the track representation and be reused in future identity comparisons. A transient unreliable branch can thus distort the current decision; even when the assignment is recovered by motion, geometry, or another reliable source, its degraded descriptor may still enter appearance memory and influence subsequent comparisons. The central challenge is therefore not only how to integrate multimodal information, but also how to preserve and control its evidence after representation formation.

These two temporal roles of appearance---current association and persistent memory---call for control beyond representation construction itself. After multimodal information has been integrated, identity evidence from different sources should remain distinguishable and adjustable according to the current observation rather than becoming irrevocably entangled in a single state. Moreover, this control should persist across time: the trust assigned to evidence when it supports a current association should be consistent with the extent to which the same evidence is allowed to alter the historical track representation. We refer to this observation-conditioned trust as \emph{appearance reliability}. It is neither a probability of correct identity nor a substitute for detector confidence; rather, it characterizes how trustworthy the available appearance evidence is for identity reasoning under the current observation.

Accordingly, we propose RCA-MOT, a \emph{reliability-consistent appearance modeling} framework for RGB--infrared MOT. As conceptually illustrated in Fig.~\ref{fig:motivation}(c), RCA-MOT retains RGB-specific, infrared-specific, and fused identity representations so that modality evidence remains accessible after representation formation. A simple training-free target-scene estimator then derives branch-wise reliability from each target and its local surroundings. During association, relative reliability adjusts the three branch contributions, while absolute reliability adds caution for borderline candidates near the ReID acceptance boundary; after a match is accepted, the same branch-wise reliability controls how strongly each descriptor is written into historical appearance memory. Fig.~\ref{fig:framework_overview} summarizes the complete architecture. We deliberately use standard branch encoders, simple fusion, standard ReID supervision, training-free reliability estimation, and conventional exponential-moving-average memory regulation, keeping the formulation focused on how multimodal appearance is organized and controlled rather than on increasingly elaborate feature interaction or learned reliability estimation. Experiments on VT-MOT and UniRTL show that RCA-MOT surpasses the previous SOTA by 5.402/6.075 HOTA and 11.549/12.789 IDF1, respectively; further analyses confirm the effectiveness of appearance reliability in regulating association and historical appearance.

Our main contributions are summarized as follows:
\begin{itemize}
    \item We revisit RGB--infrared MOT from the perspective of \emph{appearance utilization} and identify a structural gap between multimodal information integration and post-representation evidence controllability across current association and persistent track states.

    \item We propose RCA-MOT, which maintains RGB-specific, infrared-specific, and fused identity representations as controllable appearance states and uses observation-level reliability as a shared control signal for association and appearance-memory update.

    \item We instantiate this framework with training-free target-scene reliability estimation, reliability-aware association, and branch-wise memory regulation. Experiments across two benchmarks and two detection sources show consistent identity-association gains and support the estimated reliability as an informative control signal.
\end{itemize}
\section{Related Work}
\label{sec:related_work}

\subsection{RGB--Infrared Multi-Object Tracking}
\label{subsec:rw_rgbir}

RGB--infrared multi-object tracking has developed primarily around how complementary visible and thermal information can be exploited for identity-preserving tracking. Earlier RGB--thermal tracking studies explored modality-specific adaptation, quality-aware fusion, challenge-aware interaction, and mutual conditioning~\cite{li2019manet,li2020cat,zhu2021fanet,lu2025dmcnet}. Large-scale benchmarks such as LasHeR and VTUAV subsequently broadened the data and evaluation conditions for multimodal tracking~\cite{li2022lasher,zhang2022vtuav}. More recent approaches extend this representation-centered line through visual prompts, search--template interaction, self-distillation, and cross-modulated attention~\cite{zhu2023vipt,hui2023tbsi,hou2024sdstrack,xiao2025caformer}. In RGB--infrared MOT, PFTrack introduces progressive visible--thermal interaction and temporal modeling, UniS-MOT integrates detection and long-term association within a unified pipeline, and MCTrack further strengthens cross-modality spatial--temporal feature interaction~\cite{pftrack,unirtl,mctrack}. Despite substantial differences in implementation, the dominant objective remains similar: improving how RGB and infrared information is integrated when the identity representation is formed. Comparatively less attention has been paid to what happens \emph{after} representation formation---whether modality-specific and fused identity evidence should remain separately controllable, and whether such control should extend from current association to the evolution of trajectory appearance. Our work focuses on this post-representation stage, studying how already-available multimodal identity evidence should be organized, interpreted, and carried forward in online tracking.

\begin{figure*}[!t]
\centering
\paperfigure{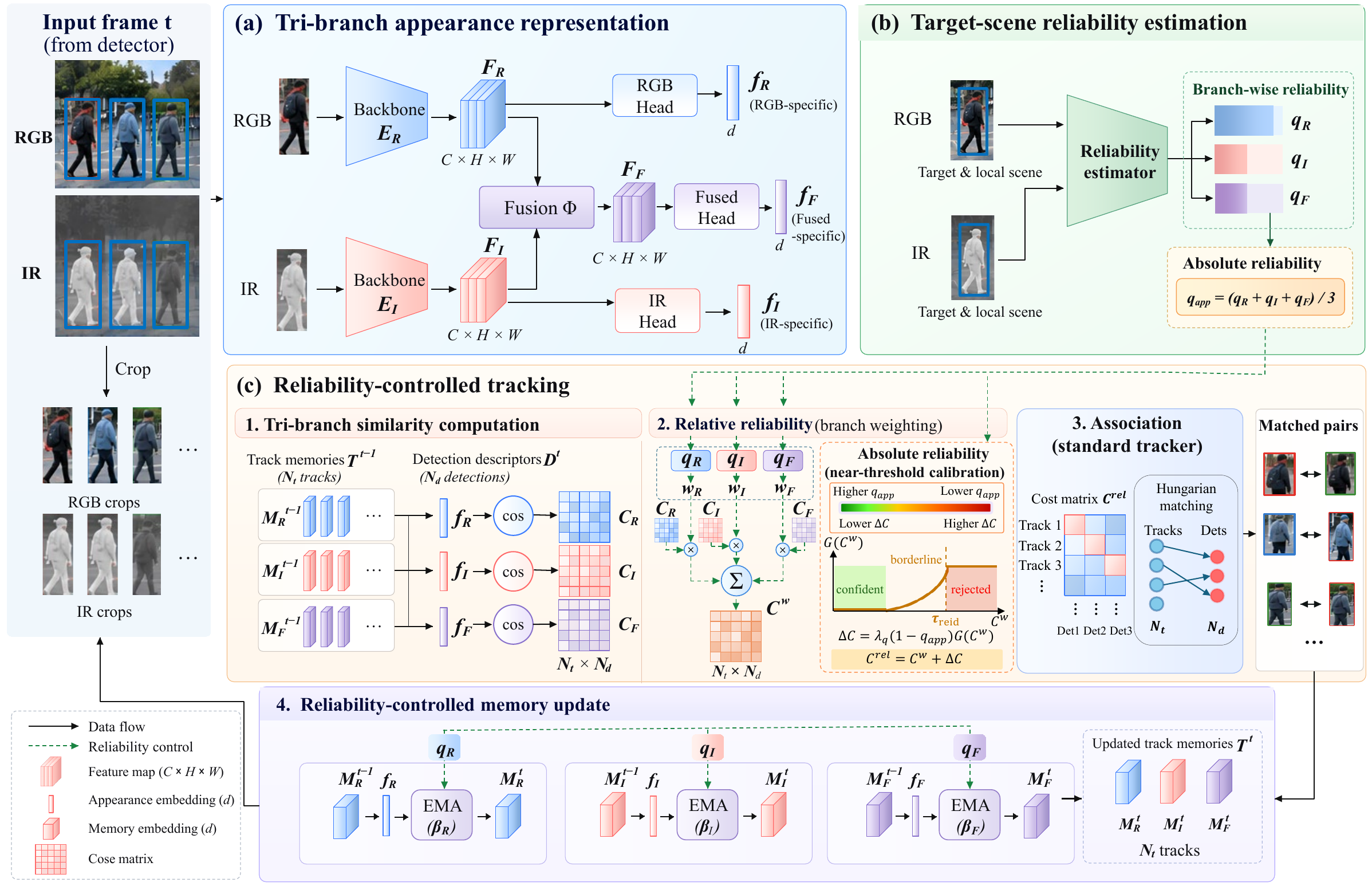}{0.86\textwidth}{7.2cm}{Fig. 2 placeholder: complete reliability-consistent tracking framework.}
\caption{Overview of RCA-MOT. Target-scene reliability regulates branch evidence and near-threshold calibration during association, and controls how strongly accepted descriptors update branch-wise memories.}
\label{fig:framework_overview}
\end{figure*}

\subsection{Appearance Modeling in Online Multi-Object Tracking}
\label{subsec:rw_appearance}

Appearance has evolved from an auxiliary matching cue into a persistent source of identity information in online MOT. SORT and DeepSORT established the tracking-by-detection paradigm and the value of deep ReID embeddings~\cite{bewley2016sort,deepsort}, while later joint and transformer-based trackers increasingly coupled localization, identity reasoning, and temporal state propagation~\cite{zhang2021fairmot,zhou2020centertrack,wu2021trades,sun2020transtrack,meinhardt2022trackformer,zeng2022motr,zhang2023motrv2,gao2023memotr}. Subsequent trackers further improved association by combining appearance with motion, detection confidence, observation-centric modeling, and transferable identity cues~\cite{du2022strongsort,aharon2022botsort,maggiolino2023deepocsort,hybridsort,bytetrack,ocsort,pang2021qdtrack,seidenschwarz2023ghost}. Progress in ReID has simultaneously strengthened the identity metric through triplet learning, BNNeck-style baselines, Circle Loss, and transformer-based representations~\cite{hermans2017triplet,luo2019bagtricks,sun2020circleloss,he2021transreid}. Collectively, these studies show that appearance is not merely a stand-alone descriptor: it participates in both the current track--detection comparison and the historical representation reused by future frames. Existing work has substantially improved association and appearance memory, but they are commonly treated as adjacent yet separate design problems. In RGB--infrared tracking, where modality quality can change independently, the trust assigned to current evidence and the extent to which the same evidence should alter persistent identity memory are naturally coupled. This motivates explicitly connecting appearance use with appearance accumulation.

\subsection{Reliability-Aware Appearance Processing}
\label{subsec:rw_reliability}

Unequal observation quality has been modeled in several related areas, and appearance reliability itself is not a new concept. In MOT, confidence-aware association uses detector scores to regulate how observations participate in tracking~\cite{bytetrack}; in ReID, quality-aware set modeling adjusts the contribution of individual observations~\cite{liu2017qan}; and in RGB--thermal single-object tracking, quality-aware methods adapt fusion or modality weighting to the current observation~\cite{qatrgbt}. Multimodal learning has likewise studied modality imbalance through gradient modulation and modality dropout~\cite{wang2020multimodalhard,peng2022ogmge,neverova2016moddrop}. These methods establish the value of reliability information, but typically apply it to a local operation such as detection selection, feature fusion, or representation aggregation. RGB--infrared MOT introduces a different system structure: appearance must first support pairwise identity association and may then become part of a persistent trajectory representation. Quality-aware fusion alone therefore does not determine how unreliable branch evidence should affect association, while association-only calibration does not determine how strongly an accepted observation should modify appearance memory. Our distinction lies not in proposing another quality score, but in assigning reliability a shared system role: RGB-specific, infrared-specific, and fused evidence remain controllable, and the same observation-level signal regulates both current association and persistent appearance update.
\section{Method}
\label{sec:method}

RCA-MOT modifies only the appearance-related components within a standard tracking-by-detection pipeline, as summarized in Fig.~\ref{fig:framework_overview}. Detection boxes and scores, motion prediction, geometric matching, assignment, and track management remain unchanged. The following sections present the controllable tri-branch appearance representation, target-scene reliability for association calibration, and reliability-guided appearance memory.

\begin{figure*}[!t]
\centering
\paperfigure{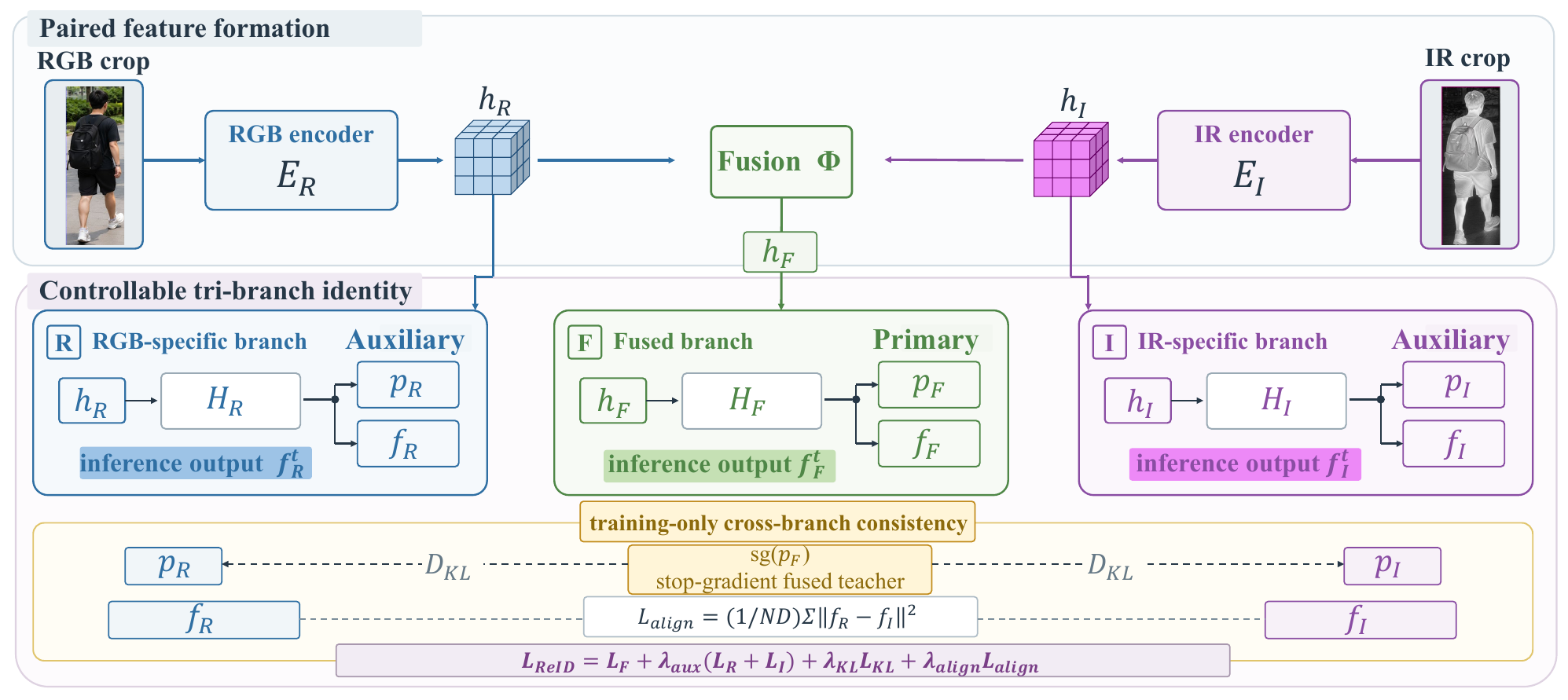}{0.8\textwidth}{6.4cm}{Fig. 3 placeholder: tri-branch representation and training objective.}
\caption{Controllable tri-branch representation and training objective. Modality-specific encoders and a simple fusion branch retain three usable descriptors; branch-wise supervision, prediction consistency, and feature alignment jointly constrain their identity geometry.}
\label{fig:tri_branch}
\end{figure*}

\subsection{Controllable Tri-Branch Appearance Representation}
\label{subsec:tri}

Existing RGB--infrared MOT systems such as PFTrack and UniS-MOT~\cite{pftrack,unirtl} mainly form a unified multimodal representation through cross-modal interaction or feature fusion and use the fused identity feature for tracking. Such a design integrates complementary information effectively, but modality-specific identity evidence is not retained as separately controllable tracking information after fusion. RCA-MOT therefore preserves RGB- and infrared-specific descriptors alongside the fused descriptor, allowing the three sources of appearance evidence to be compared and maintained separately during association and memory update (Fig.~\ref{fig:tri_branch}).

For detection $j$ at frame $t$, let $\bm z_{j,\mathrm R}^{t}$ and $\bm z_{j,\mathrm I}^{t}$ be synchronized target crops. Independently parameterized encoders extract modality-specific features, which are also fused:
\begin{equation}
\begin{aligned}
\bm h_{j,\mathrm R}^{t}
&=
E_{\mathrm R}
\left(
\bm z_{j,\mathrm R}^{t}
\right),
\\
\bm h_{j,\mathrm I}^{t}
&=
E_{\mathrm I}
\left(
\bm z_{j,\mathrm I}^{t}
\right),
\\
\bm h_{j,\mathrm F}^{t}
&=
\Phi
\left(
\left[
\bm h_{j,\mathrm R}^{t}
\Vert
\bm h_{j,\mathrm I}^{t}
\right]
\right),
\end{aligned}
\label{eq:method_tri_representation}
\end{equation}
Here, $\Vert$ denotes channel-wise concatenation, and $\Phi$ comprises a $1\times1$ convolution, batch normalization, and ReLU. Branch-specific embedding heads produce $\ell_2$-normalized identity descriptors $\bm f_{j,\mathrm R}^{t}$, $\bm f_{j,\mathrm I}^{t}$, and $\bm f_{j,\mathrm F}^{t}$. We use standard encoders and simple fusion to focus on controllable appearance organization rather than feature-extractor complexity.

The three branches remain accessible throughout tracking: each provides a current descriptor and maintains a corresponding historical reference. This organization allows observation quality to regulate both how strongly each branch contributes to current matching and how strongly its accepted descriptor is accumulated into the track representation, directly connecting appearance representation with its temporal use.

\textbf{Branch-wise discrimination.}
Features trained only to assist fusion need not support independent identity matching. However, the subsequent association explicitly compares the RGB and infrared descriptors with their corresponding track memories, so each modality-specific branch must directly learn the identity metric used at inference time. We therefore supervise every branch with identity classification and hard-mining triplet losses:
\begin{equation}
\mathcal L_b
=
\mathcal L_{\mathrm{id}}^{b}
+
\mathcal L_{\mathrm{tri}}^{b},
\qquad
b\in\{\mathrm R,\mathrm I,\mathrm F\}.
\label{eq:method_branch_loss}
\end{equation}
Classification establishes identity-level discrimination, while triplet supervision directly shapes the embedding distances used for ReID matching. The fused branch receives the primary weight and the RGB/infrared branches smaller auxiliary weights; ``auxiliary'' refers only to their relative training weights, as both modality-specific descriptors are explicitly used during tracking. This supervision converts the RGB and infrared outputs from fusion-oriented intermediate features into identity representations that can be independently matched and maintained.

\begin{figure*}[!t]
\centering
\paperfigure{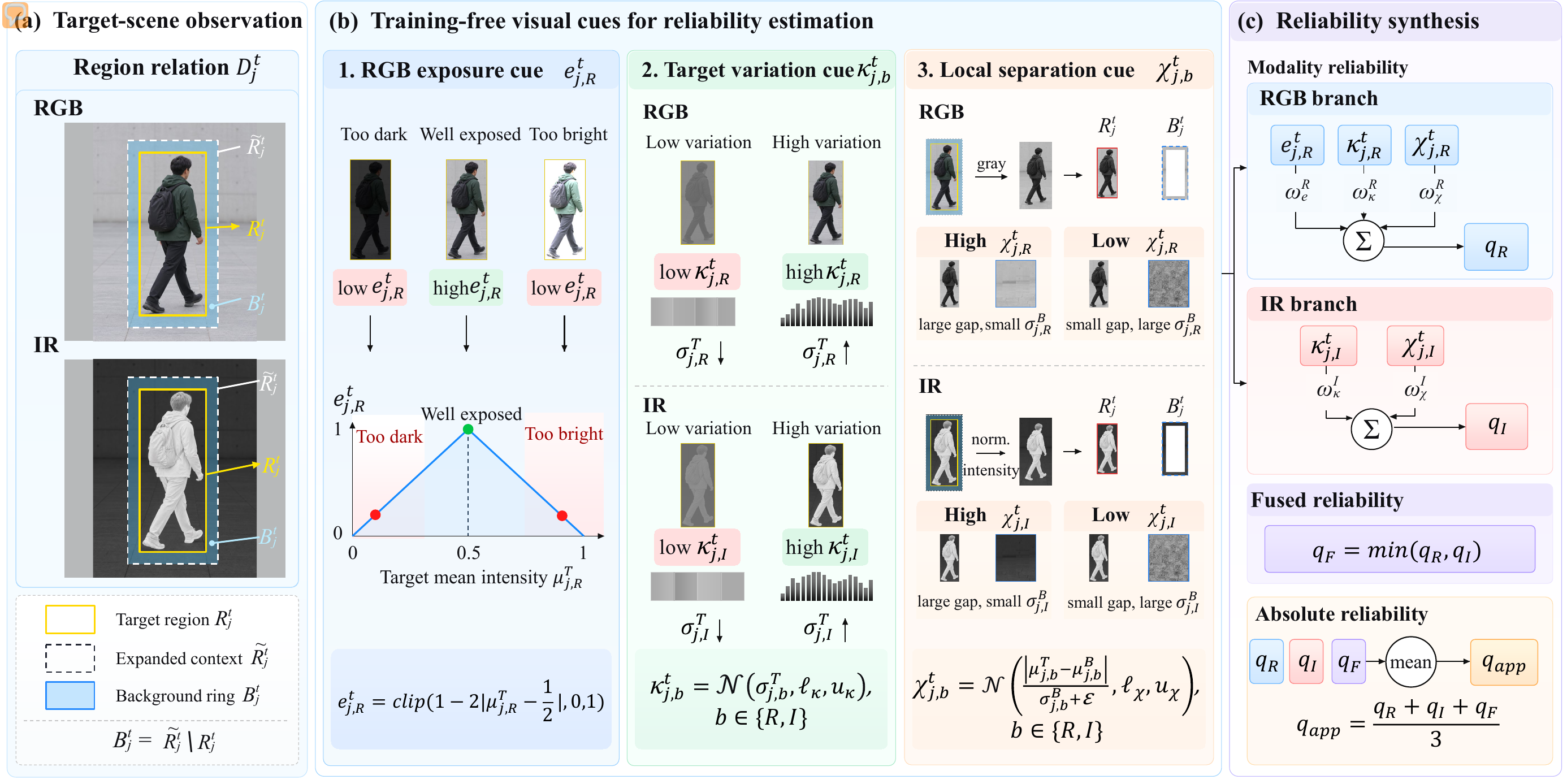}{0.8\textwidth}{6.4cm}{Fig. 4 placeholder: target-scene reliability estimator.}
\caption{Target-scene branch-reliability estimation. RGB uses exposure, target variation, and local separability; infrared uses the latter two cues. Fused reliability is conservatively defined as $q_{\mathrm F}=\min(q_{\mathrm R},q_{\mathrm I})$.}
\label{fig:reliability_estimator}
\end{figure*}

\textbf{Cross-branch consistency.}
Independent identity supervision ensures that each branch can support ReID, but the three branch distances are subsequently combined within the same association decision. Individually discriminative branches may still differ in class relations or contain modality-dependent geometric offsets unrelated to identity; without additional constraints, reliability weighting would therefore respond to both observation reliability and irrelevant cross-branch mismatch. We first reduce this discrepancy in prediction space. Let $\bm p_b$ denote the class distribution of branch $b$; the fused prediction provides a joint identity reference for the modality-specific branches through distillation~\cite{hinton2015distilling}:
\begin{equation}
\mathcal L_{\mathrm{KL}}
=
D_{\mathrm{KL}}
\left(
\operatorname{sg}
[\bm p_{\mathrm F}]
\Vert
\bm p_{\mathrm R}
\right)
+
D_{\mathrm{KL}}
\left(
\operatorname{sg}
[\bm p_{\mathrm F}]
\Vert
\bm p_{\mathrm I}
\right),
\label{eq:method_kl}
\end{equation}
Here, $\operatorname{sg}$ denotes stop-gradient, preventing this term from changing the fused prediction through the auxiliary branches. Since prediction consistency does not directly align ReID geometry, we also constrain synchronized RGB and infrared embeddings:
\begin{equation}
\mathcal L_{\mathrm{align}}
=
\frac{1}{ND}
\sum_{n=1}^{N}
\left\|
\bm f_{n,\mathrm R}
-
\bm f_{n,\mathrm I}
\right\|_2^2,
\label{eq:method_alignment}
\end{equation}
Here, $N$ and $D$ are the mini-batch size and embedding dimension. This term reduces unnecessary cross-modal offsets without forcing the two encoders to learn identical mappings or removing modality-specific information. Prediction consistency transfers the fused branch's relative identity structure, while alignment encourages more comparable embedding geometry for subsequent joint association; neither replaces the branch-wise discrimination objective. The complete objective is
\begin{equation}
\mathcal L_{\mathrm{ReID}}
=
\mathcal L_{\mathrm F}
+
\lambda_{\mathrm{aux}}
\left(
\mathcal L_{\mathrm R}
+
\mathcal L_{\mathrm I}
\right)
+
\lambda_{\mathrm{KL}}
\mathcal L_{\mathrm{KL}}
+
\lambda_{\mathrm{align}}
\mathcal L_{\mathrm{align}},
\label{eq:method_reid_objective}
\end{equation}
with weights specified in Section~\ref{subsec:experimental_setup}. Together, branch supervision and cross-branch constraints provide three usable identity representations with more consistent semantics and geometry. The next step determines how much each representation should be trusted for the current observation.

\subsection{Target-Scene Reliability and Association Calibration}
\label{subsec:association}

The three branches are separately accessible, but their usefulness changes with local illumination and thermal conditions. Detector confidence primarily reflects objectness: a confident detection can still be underexposed, nearly uniform, or poorly separated from its background. We therefore estimate detection-specific appearance reliability from the target and its surroundings.

\textbf{Local target-scene reliability.}
Target-internal variation alone cannot describe local separability, while whole-frame statistics may overlook spatially varying conditions. We expand the target box $\mathcal R_j^t$ by factor $\eta$ to obtain $\widetilde{\mathcal R}_j^t$, and use the surrounding ring $\mathcal B_j^t=\widetilde{\mathcal R}_j^t\setminus\mathcal R_j^t$ as the background reference (Fig.~\ref{fig:reliability_estimator}). This captures both information inside the target and its contrast with the local scene.

On normalized grayscale observations, let $\mu_{j,b}^{T}$, $\sigma_{j,b}^{T}$ and
$\mu_{j,b}^{B}$, $\sigma_{j,b}^{B}$ denote target and background means and standard
deviations for $b\in\{\mathrm R,\mathrm I\}$. Using
$\mathcal N(x;\ell,u)=\operatorname{clip}((x-\ell)/(u-\ell),0,1)$, where $\ell$ and $u$
denote the lower and upper normalization bounds, respectively, we construct
\begin{equation}
\begin{aligned}
e_{j,\mathrm R}^{t}
&=
\operatorname{clip}
\left(
1-
2\left|
\mu_{j,\mathrm R}^{T}-\frac12
\right|,
0,1
\right),
\\
\kappa_{j,b}^{t}
&=
\mathcal N
\left(
\sigma_{j,b}^{T};
\ell_{\kappa},
 u_{\kappa}
\right),
\\
\chi_{j,b}^{t}
&=
\mathcal N
\left(
\frac{
\left|
\mu_{j,b}^{T}-\mu_{j,b}^{B}
\right|
}{
\sigma_{j,b}^{B}+\epsilon
};
\ell_{\chi},
 u_{\chi}
\right).
\end{aligned}
\label{eq:method_reliability_components}
\end{equation}
The exposure cue $e_{\mathrm R}$ describes the overall exposure condition of the RGB target region. It is high for intermediate target intensities and decreases toward extremely dark or bright observations, identifying cases in which visible appearance becomes less usable under extreme intensity conditions. The target-variation cue $\kappa_b$ characterizes the amount of internal intensity variation within the target. A nearly uniform target has a small standard deviation and provides less local appearance structure, whereas richer internal variation yields a larger $\kappa_b$.  The local-separation cue $\chi_b$ complements this target-internal measure by describing how strongly the target differs from its immediate surroundings relative to background fluctuations. A larger target--background mean gap and a more stable local background provide stronger separation evidence, whereas the same mean gap is less informative when the surrounding background itself varies substantially. Thus, $\kappa_b$ and $\chi_b$ characterize complementary aspects of appearance usability: target-internal structure and local target--scene separation, respectively, while $e_{\mathrm R}$ additionally captures extreme RGB exposure. These cues are training-free proxies for appearance usability rather than learned semantic confidence in identity.

\textbf{Modality-specific and absolute reliability.}
RGB reliability combines all three cues:
\begin{equation}
q_{j,\mathrm R}^{t}
=
\operatorname{clip}
\left(
\omega_{e}^{\mathrm R}e_{j,\mathrm R}^{t}
+
\omega_{\kappa}^{\mathrm R}\kappa_{j,\mathrm R}^{t}
+
\omega_{\chi}^{\mathrm R}\chi_{j,\mathrm R}^{t},
0,1
\right).
\label{eq:method_rgb_reliability}
\end{equation}
Infrared intensity depends on scene temperature and sensor response, so deviation from mid-gray lacks the same exposure interpretation. Infrared therefore uses target variation and local thermal separation:
\begin{equation}
q_{j,\mathrm I}^{t}
=
\operatorname{clip}
\left(
\omega_{\kappa}^{\mathrm I}\kappa_{j,\mathrm I}^{t}
+
\omega_{\chi}^{\mathrm I}\chi_{j,\mathrm I}^{t},
0,1
\right).
\label{eq:method_ir_reliability}
\end{equation}
Since fusion depends on both inputs, we conservatively define fused reliability and aggregate absolute appearance usability as
\begin{equation}
\begin{aligned}
q_{j,\mathrm F}^{t}
&=
\min
\left(
q_{j,\mathrm R}^{t},
q_{j,\mathrm I}^{t}
\right),
\\
q_{j,\mathrm{app}}^{t}
&=
\frac{
q_{j,\mathrm R}^{t}
+
q_{j,\mathrm I}^{t}
+
q_{j,\mathrm F}^{t}
}{3}.
\end{aligned}
\label{eq:method_reliability_scores}
\end{equation}
The minimum avoids assuming that fusion is unaffected by a degraded input. The aggregate $q_{j,\mathrm{app}}^t$ expresses overall appearance usability, not a calibrated probability of correct matching.

\textbf{Relative reliability weighting.}
Let $C_{k,j,b}^{t}$ be the cosine distance between detection $j$ and track $k$'s branch memory. Reliability is observation-specific, whereas this distance measures pair-specific identity compatibility. Thus, a reliable observation is not automatically a match to any particular track: its branch distances still determine identity compatibility. We use reliability to regulate their contributions:
\begin{equation}
\begin{aligned}
w_{j,b}^{t}
&=
\frac{
q_{j,b}^{t}
}{
\sum_{c\in\{\mathrm R,\mathrm I,\mathrm F\}}
q_{j,c}^{t}
},
\\
C_{k,j}^{w,t}
&=
\sum_{b\in\{\mathrm R,\mathrm I,\mathrm F\}}
w_{j,b}^{t}
C_{k,j,b}^{t}.
\end{aligned}
\label{eq:method_relative_cost}
\end{equation}
If all reliabilities are zero, uniform weighting is used. Otherwise, clearer branches contribute more, with soft weights accommodating imperfect reliability estimates. However, normalized weights always sum to one: they indicate which branch is relatively preferable but cannot express that all branches are weak. Absolute reliability supplies this missing information.

\textbf{Absolute near-threshold calibration.}
Penalizing every low-quality observation equally could damage a clear low-cost match, while an already rejected pair needs no correction. We therefore focus on weak observations whose weighted cost lies just inside the ReID acceptance boundary:
\begin{equation}
\begin{aligned}
G(C)
&=
\mathbf 1
\left[
C<\tau_{\mathrm{reid}}
\right]
\exp
\left(
-\frac{
\tau_{\mathrm{reid}}-C
}{
\tau_p
}
\right),
\\
C_{k,j}^{\mathrm{rel},t}
&=
C_{k,j}^{w,t}
+
\lambda_q
\left(
1-q_{j,\mathrm{app}}^{t}
\right)
G
\left(
C_{k,j}^{w,t}
\right).
\end{aligned}
\label{eq:method_absolute_calibration}
\end{equation}
Here, $\tau_{\mathrm{reid}}$ is the original acceptance threshold, $\tau_p$ controls the boundary-window scale, and $\lambda_q$ sets the correction magnitude. The correction is small far below the threshold, increases toward it from below, and is zero outside it. Being nonnegative, it can make a borderline pair harder to accept but never more attractive. Thus, relative reliability balances branch contributions, while absolute reliability adds caution when borderline appearance evidence is weak.

The calibrated $C_{k,j}^{\mathrm{rel},t}$ replaces only the ReID term in the existing association stage. This controls the current identity decision; the next section controls how accepted evidence affects future comparisons.

\subsection{Reliability-Guided Appearance Memory}
\label{subsec:memory}

A correct match does not imply that every branch is suitable for memory update. For example, motion and clear infrared appearance can recover a target whose RGB crop is underexposed, yet fixed EMA still writes its degraded RGB descriptor into history. We therefore retain the assignment and regulate each branch's write strength, allowing reliable evidence to update while limiting contamination from degraded observations.

\textbf{Reliability-controlled memory update.}
Each track maintains normalized branch memories $\bm m_{k,\mathrm R}$, $\bm m_{k,\mathrm I}$, and $\bm m_{k,\mathrm F}$. For detection $j$ matched to track $k$ at frame $t$, reliability determines the current descriptor's contribution:
\begin{equation}
\beta_{j,b}^{t}
=
(1-\alpha)
\operatorname{clip}
\left(
\gamma q_{j,b}^{t},
0,1
\right),
\qquad
b\in\{\mathrm R,\mathrm I,\mathrm F\},
\label{eq:method_memory_injection}
\end{equation}
Here, $\alpha$ is the base EMA historical momentum and $\gamma$ scales reliability modulation. Reliability does not directly decay or rescale the existing memory; it regulates only how strongly the current descriptor is admitted into the historical representation. The memory update is
\begin{equation}
\bm m_{k,b}^{t}
=
\operatorname{Norm}
\left[
\left(
1-\beta_{j,b}^{t}
\right)
\bm m_{k,b}^{t-1}
+
\beta_{j,b}^{t}
\bm f_{j,b}^{t}
\right],
\label{eq:method_memory_update}
\end{equation}
where $\operatorname{Norm}$ denotes $\ell_2$ normalization. Since $0\leq\beta_{j,b}^{t}\leq1-\alpha$, the gate never admits more current evidence than the base EMA. Lower reliability smoothly reduces the write strength, preserving more history; higher reliability permits a larger contribution within this bound. Branches update independently, so a clear modality can adapt while a degraded one is restrained.

Unlike a larger fixed EMA momentum, this write strength varies across observations and branches. A fixed momentum slows all updates equally, whereas reliability distinguishes useful current evidence from a temporarily degraded branch. A newly initialized branch uses its first valid descriptor directly; reliability-controlled interpolation applies to subsequent matched observations.

\textbf{Reliability-consistent association and memory.}
The same reliability signal governs two distinct operations: association calibration limits weak evidence in the current decision, while memory gating limits its influence on future references. Let $\Delta C_{k,j}^{t}=C_{k,j}^{\mathrm{rel},t}-C_{k,j}^{w,t}$. With the remaining quantities fixed, wherever the derivatives exist,
\begin{equation}
\frac{\partial \Delta C_{k,j}^{t}}
{\partial q_{j,\mathrm{app}}^{t}}
\le 0,
\qquad
\frac{\partial \beta_{j,b}^{t}}
{\partial q_{j,b}^{t}}
\ge 0.
\label{eq:method_reliability_consistency}
\end{equation}
Thus, decreasing absolute reliability cannot reduce the calibration penalty, and decreasing branch reliability cannot increase memory injection. Association-only control would still admit degraded features after a match recovered by other cues; memory-only control could not undo an already corrupted assignment. Applying reliability at both stages limits the immediate influence and subsequent propagation of unreliable appearance evidence.

This shared control connects the trust assigned to an observation now with the influence it retains later. Relative weighting, absolute calibration, and branch-wise updates govern complementary uses of the same evidence: its contribution across branches, its acceptance near the matching boundary, and its admission into trajectory history.
\section{Experiments}
\label{sec:experiments}

\subsection{Experimental Setup}
\label{subsec:experimental_setup}

\begin{table*}[!t]
\centering
\caption{VT-MOT results under Protocol~I. Parentheses identify the fixed offline detections used by RCA-MOT; cross-method results are benchmark-level comparisons.}
\label{tab:vtmot_sota}
\small
\setlength{\tabcolsep}{9pt}
\begin{tabular}{@{}lccccc@{}}
\toprule
Method & HOTA $\uparrow$ & DetA $\uparrow$ & MOTA $\uparrow$ & MOTP $\uparrow$ & IDF1 $\uparrow$ \\
\midrule
FairMOT~\cite{zhang2021fairmot}        & 37.350 & 34.628 & 37.266 & 72.525 & 45.795 \\
CenterTrack~\cite{zhou2020centertrack} & 39.045 & 38.104 & 30.585 & 72.874 & 44.420 \\
TraDeS~\cite{wu2021trades}             & 38.319 & 36.370 & 34.632 & 72.295 & 47.008 \\
TransTrack~\cite{sun2020transtrack}    & 38.000 & 35.711 & 36.156 & 73.823 & 43.567 \\
ByteTrack~\cite{bytetrack}             & 38.393 & 32.122 & 33.151 & 73.483 & 45.757 \\
OC-SORT~\cite{ocsort}                  & 31.479 & 25.244 & 28.948 & 73.150 & 38.086 \\
MixSort-OC~\cite{cui2023sportsmot}     & 39.090 & 33.109 & 31.330 & 73.632 & 45.799 \\
MixSort-Byte~\cite{cui2023sportsmot}   & 39.575 & 34.806 & 31.593 & 73.049 & 46.367 \\
PID-MOT~\cite{lv2024pidmot}            & 35.621 & 33.245 & 33.333 & 71.794 & 42.430 \\
Hybrid-SORT~\cite{hybridsort}          & 39.485 & 34.619 & 31.074 & 72.840 & 46.310 \\
PFTrack~\cite{pftrack}                 & 41.068 & \textbf{41.631} & 43.088 & 73.949 & 47.254 \\
UniS-MOT~\cite{unirtl}                 & 41.730 & 41.270 & 32.650 & 72.740 & 47.310 \\
\midrule
RCA-MOT (PFTrack det.)   & 45.042 & 39.143 & 41.791 & 74.069 & 57.108 \\
RCA-MOT (YOLO-RGBT det.) & \textbf{47.132} & 40.921 & \textbf{47.029} & \textbf{76.537} & \textbf{58.859} \\
\bottomrule
\end{tabular}
\end{table*}

\begin{table}[!t]
\centering
\caption{UniRTL test results. PFTrack and UniS-MOT are reproduced on the same 30/20 split.}
\label{tab:unirtl_sota}
\small
\setlength{\tabcolsep}{2.8pt}
\resizebox{\columnwidth}{!}{%
\begin{tabular}{@{}lccccc@{}}
\toprule
Method & HOTA $\uparrow$ & DetA $\uparrow$ & MOTA $\uparrow$ & MOTP $\uparrow$ & IDF1 $\uparrow$ \\
\midrule
PFTrack$^{\dagger}$~\cite{pftrack} & 46.568 & \textbf{55.043} & 66.946 & 77.732 & 52.911 \\
UniS-MOT$^{\dagger}$~\cite{unirtl} & 49.062 & 50.678 & 50.577 & 78.905 & 56.816 \\
RCA-MOT (PFTrack det.)      & \textbf{55.137} & \textbf{55.043} & \textbf{68.028} & 78.140 & \textbf{69.605} \\
RCA-MOT (YOLO-RGBT det.)    & 49.384 & 48.484 & 54.382 & \textbf{79.033} & 61.225 \\
\bottomrule
\end{tabular}
}
\vspace{1mm}
\parbox{0.97\columnwidth}{\footnotesize $^{\dagger}$Reproduced using the released UniRTL split.}
\end{table}

\textbf{Datasets and evaluation.}
We evaluate RCA-MOT on VT-MOT~\cite{pftrack} under Protocol~I (462 training and 120 test sequences) and the RGB--infrared MOT subset of UniRTL~\cite{unirtl}. For UniRTL, the 20 test sequences are excluded from the released 50-sequence training list, leaving 30 sequences for training. Benchmark comparisons report HOTA/DetA~\cite{luiten2021hota}, MOTA/MOTP~\cite{bernardin2008clearmot}, and IDF1~\cite{ristani2016idmetrics}, following standard MOT evaluation~\cite{dendorfer2021motchallenge}. Controlled comparisons retain complete TrackEval outputs and report HOTA, DetA, AssA, and IDF1.

\textbf{Tracking and detections.}
We use fixed offline detections from PFTrack and an RGB--thermal YOLO detector (YOLO-RGBT). Within each dataset--detector setting, all variants receive identical boxes and scores and share Kalman-filter prediction, geometric matching, assignment, and track management; only appearance representation, ReID cost, and memory update vary. In the benchmark tables, ``PFTrack'' denotes its complete pipeline and ``PFTrack det.'' its detections used as fixed input. Cross-method benchmark results may differ in detectors, training data, and implementation; component effects are assessed through the controlled comparisons.

\textbf{Implementation details.}
We train a separate ReID model per benchmark using $384\times128$ synchronized crops, independently parameterized ResNeSt-50 streams, and 2048-dimensional GeM--BNNeck descriptors. Training uses Adam for 60 epochs with learning rate $3.5\times10^{-4}$ and batch size 384. The loss weights are $\lambda_{\mathrm{aux}}=0.5$, $\lambda_{\mathrm{KL}}=1.0$, and $\lambda_{\mathrm{align}}=0.5$, with unit fused-branch weight. Tracking uses $\tau_{\mathrm{reid}}=0.6$, $\lambda_q=0.20$, $\tau_p=0.10$, $\alpha=0.8$, $\gamma=0.5$, and $\eta=1.8$. The same configuration is used across both benchmarks and detection sources.

\subsection{Overall Performance}
\label{subsec:overall_performance}

On VT-MOT (Table~\ref{tab:vtmot_sota}), RCA-MOT with PFTrack detections achieves 45.042 HOTA and 57.108 IDF1, exceeding the highest prior values listed by 3.312 and 9.798 points, respectively. YOLO-RGBT detections further raise these scores to 47.132 and 58.859. The gains are concentrated in identity-oriented metrics, without corresponding increases in DetA and MOTA under PFTrack detections.

\begin{table*}[!t]
\centering
\caption{Controlled evaluation of RCA-MOT. Full denotes the complete framework; all comparisons within each setting use identical offline detections. In (b), results are reported as Base/Full/$\Delta$.}
\label{tab:framework_evaluation}
\footnotesize
\setlength{\tabcolsep}{3.5pt}
\renewcommand{\arraystretch}{1.05}

\begin{tabular*}{0.98\textwidth}
{@{\extracolsep{\fill}}lcccccccc@{}}
\toprule
\multicolumn{9}{c}{\textbf{(a) Core components with PFTrack detections}} \\
\midrule
\multirow{4}{*}{Variant}
& \multicolumn{4}{c}{VT-MOT}
& \multicolumn{4}{c}{UniRTL} \\
\cmidrule(lr){2-5}\cmidrule(lr){6-9}
& HOTA $\uparrow$
& DetA $\uparrow$
& AssA $\uparrow$
& IDF1 $\uparrow$
& HOTA $\uparrow$
& DetA $\uparrow$
& AssA $\uparrow$
& IDF1 $\uparrow$ \\
\midrule
Baseline
& 40.896 & 39.410 & 43.672 & 49.391
& 51.126 & 55.022 & 47.756 & 63.527 \\
Tri
& 43.076 & 39.291 & 48.550 & 53.376
& 52.479 & 55.032 & 50.272 & 65.430 \\
Tri+Gate
& 43.344 & 38.639 & 49.842 & 54.438
& 54.563 & 54.863 & 54.526 & 69.084 \\
Tri+Cost
& 43.534 & 39.296 & 49.539 & 54.222
& 53.791 & 55.138 & 52.708 & 67.684 \\
Full
& \textbf{45.042}
& 39.143
& \textbf{53.055}
& \textbf{57.108}
& \textbf{55.137}
& 55.043
& \textbf{55.470}
& \textbf{69.605} \\
\bottomrule
\end{tabular*}

\vspace{1pt}

\begin{tabular*}{0.98\textwidth}
{@{\extracolsep{\fill}}lcccc@{}}
\multicolumn{5}{c}{\textbf{(b) Consistency across datasets and detection sources}} \\
\midrule
\multirow{1}{*}{Setting}
& \shortstack{HOTA\\(Base/Full/$\Delta$)}
& \shortstack{DetA\\(Base/Full/$\Delta$)}
& \shortstack{AssA\\(Base/Full/$\Delta$)}
& \shortstack{IDF1\\(Base/Full/$\Delta$)} \\
\midrule
VT-MOT--PFTrack
& $40.896/45.042/+4.146$
& $39.410/39.143/-0.267$
& $43.672/53.055/+9.383$
& $49.391/57.108/+7.717$ \\
VT-MOT--YOLO-RGBT
& $43.547/47.132/+3.585$
& $41.168/40.921/-0.247$
& $47.699/55.903/+8.204$
& $52.475/58.859/+6.384$ \\
UniRTL--PFTrack
& $51.126/55.137/+4.011$
& $55.022/55.043/+0.021$
& $47.756/55.470/+7.714$
& $63.527/69.605/+6.078$ \\
UniRTL--YOLO-RGBT
& $44.813/49.384/+4.571$
& $48.661/48.484/-0.177$
& $41.553/50.633/+9.080$
& $55.427/61.225/+5.798$ \\
\bottomrule
\end{tabular*}
\end{table*}

\begin{table}[!t]
\centering
\caption{Association--memory analysis on VT-MOT with PFTrack detections. ''Abs.'' denotes absolute near-threshold calibration.}
\label{tab:association_memory_analysis}
\small
\setlength{\tabcolsep}{4.0pt}
\begin{tabular}{@{}llccc@{}}
\toprule
Association & Memory & HOTA & AssA & IDF1 \\
\midrule
Fused          & $\alpha=0.8$ & 43.076 & 48.550 & 53.376 \\
Uniform        & $\alpha=0.8$ & 43.182 & 48.702 & 53.558 \\
Relative       & $\alpha=0.8$ & 43.144 & 48.630 & 53.491 \\
Uniform+Abs.   & $\alpha=0.8$ & 43.401 & 49.238 & 53.979 \\
Relative+Abs.  & $\alpha=0.8$ & 43.534 & 49.539 & 54.222 \\
\midrule
Uniform+Abs.   & $\alpha=0.9$ & 44.363 & 51.321 & 55.518 \\
Relative+Abs.  & $\alpha=0.9$ & 44.614 & 51.909 & 56.084 \\
\midrule
Uniform+Abs.   & Gate & 44.626 & 52.260 & 56.501 \\
Relative+Abs.  & Gate & \textbf{45.042} & \textbf{53.055} & \textbf{57.108} \\
\bottomrule
\end{tabular}
\end{table}

On UniRTL (Table~\ref{tab:unirtl_sota}), RCA-MOT with PFTrack detections reaches 55.137 HOTA and 69.605 IDF1, exceeding reproduced PFTrack by 8.569 and 16.694 points and reproduced UniS-MOT by 6.075 and 12.789 points. With YOLO-RGBT detections, it obtains 49.384 HOTA and 61.225 IDF1. These results establish benchmark-level competitiveness; the following fixed-detection experiments isolate the appearance mechanisms.

\subsection{Ablation Studies and Mechanism Analysis}
\label{subsec:controlled_ablation}

\textbf{Progressive component contribution.}
The controlled study follows the method's three stages: whether tri-branch training improves usable identity evidence, whether association and memory control each provide additional gains, and whether their combination is complementary. This separates representation learning from how the resulting descriptors are used over time.

\emph{Baseline} and \emph{Tri} share the RGB/infrared encoders, fusion module, fused embedding head, input resolution, training schedule, fused-only association, and fixed-EMA memory. Baseline trains only the fused ReID branch; Tri adds modality-specific heads, branch-wise identity/triplet supervision, and KL/alignment constraints. Both use only the fused descriptor during tracking, isolating the effect of tri-branch training from an inference-time ensemble. \emph{Tri+Gate} and \emph{Tri+Cost} separately add reliability-guided memory and reliability-aware three-branch association; \emph{Full} combines both.

Table~\ref{tab:framework_evaluation}(a) shows that Tri improves HOTA by 2.180 on VT-MOT and 1.353 on UniRTL, with higher AssA/IDF1 and nearly unchanged DetA. These gains therefore support the added training supervision. Gate and Cost each improve over Tri, and Full achieves the best HOTA, AssA, and IDF1 on both benchmarks, supporting complementary control of current association and historical appearance.

\textbf{Consistency across datasets and detection sources.}
Across the four settings in Table~\ref{tab:framework_evaluation}(b), Full improves Baseline by 3.585--4.571 HOTA, 7.714--9.383 AssA, and 5.798--7.717 IDF1 under identical detections. DetA changes only by $-0.267$ to $+0.021$. The consistent identity gains with stable detection accuracy support the effectiveness of the appearance mechanisms across datasets and detection sources.

\begin{figure*}[!t]
\centering
\paperfigure{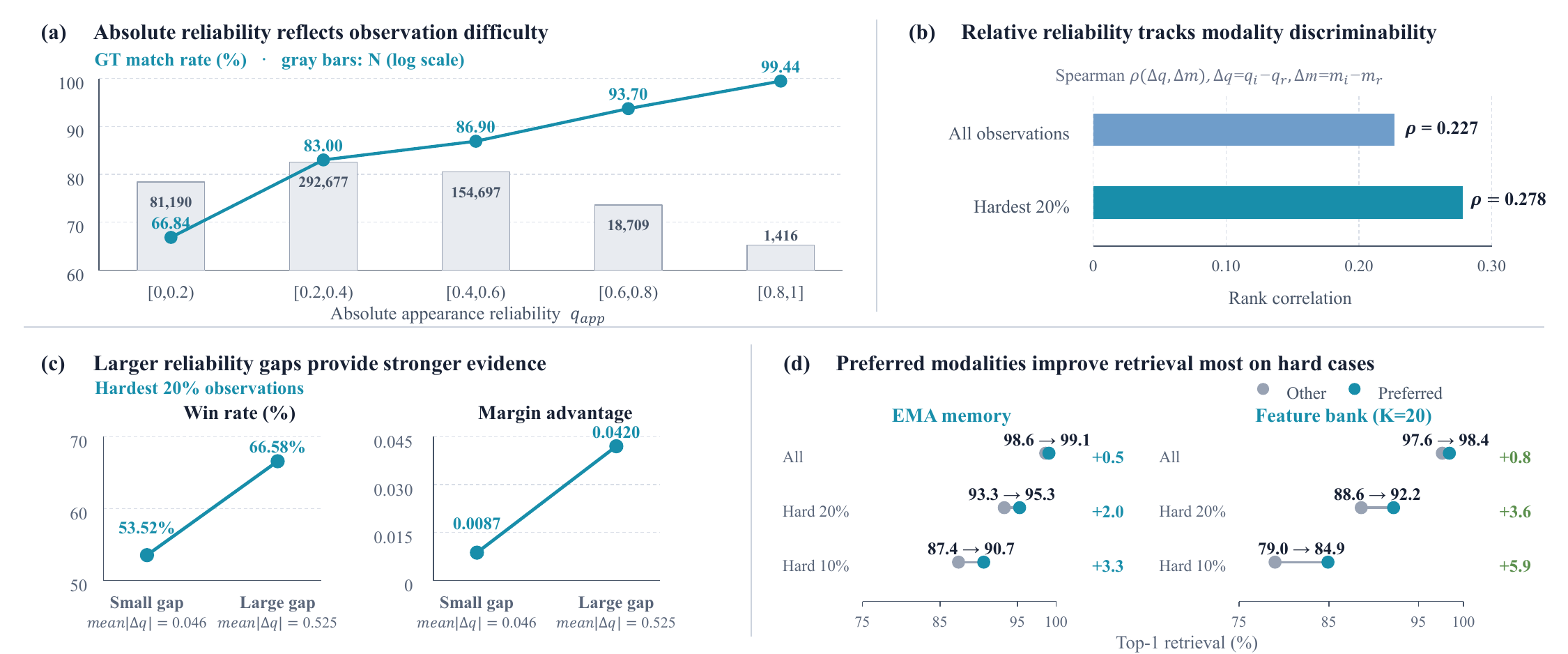}{0.95\textwidth}{5.0cm}{Fig. 5 placeholder: reliability-validity analysis.}
\caption{Validity of appearance reliability on VT-MOT. Post-hoc analyses relate absolute reliability to observation difficulty and relative RGB--infrared reliability to modality discriminability, reliability gap, and retrieval under two memory representations.}
\label{fig:reliability_validity}
\end{figure*}

\textbf{Association--memory interaction.}
Table~\ref{tab:association_memory_analysis} uses the same tri-branch ReID model on VT-MOT with PFTrack detections. Under fixed $\alpha=0.8$, Uniform slightly exceeds fused-only association, but Relative alone does not exceed Uniform. Adding absolute calibration improves Uniform and Relative by 0.219 and 0.390 HOTA, respectively; Relative+Abs. then exceeds Uniform+Abs. by 0.133 HOTA. Thus, relative weighting alone is insufficient, while absolute calibration provides the larger direct association gain in this setting.

\begin{figure*}[!t]
\centering
\paperfigure{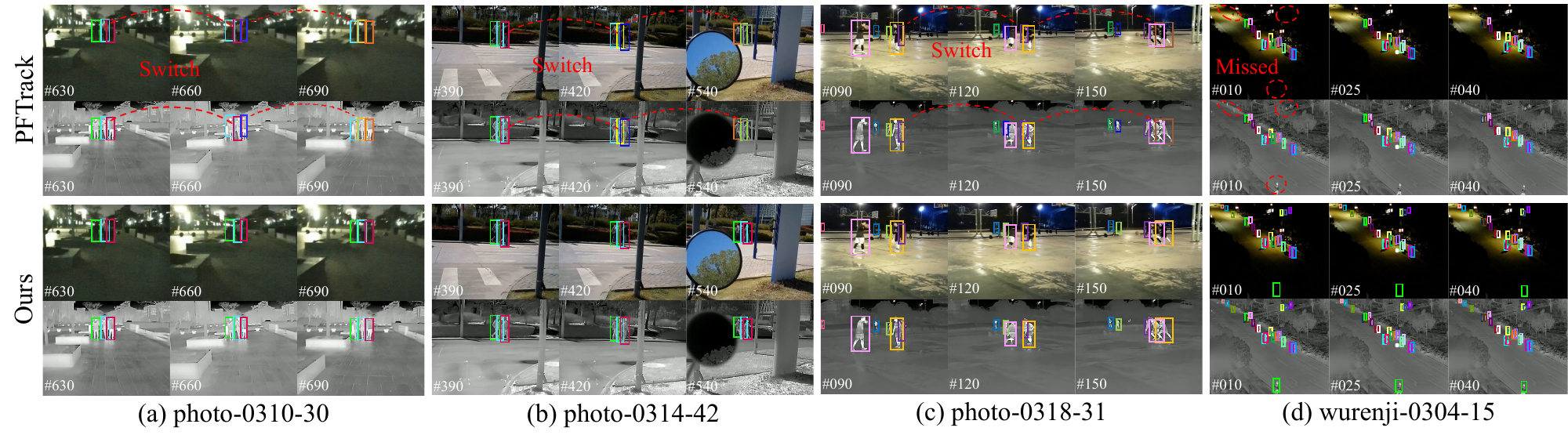}{0.98\textwidth}{5.2cm}{Fig. 6 placeholder: qualitative comparison between PFTrack and RCA-MOT with their respective detection sources.}
\caption{Qualitative comparison between PFTrack and our best-performing RCA-MOT configuration on VT-MOT. Each group presents synchronized RGB and infrared observations, with PFTrack shown above and RCA-MOT below. Panels (a)--(c) illustrate identity switches or inconsistent identity preservation, while panel (d) shows a missed or interrupted trajectory.}
\label{fig:qualitative_results}
\end{figure*}

The Relative+Abs. advantage over Uniform+Abs. increases to 0.251 HOTA with fixed $\alpha=0.9$ and 0.416 with Gate, with corresponding AssA/IDF1 gains. This pattern supports an interaction between current branch weighting and the quality of historical references. Relative reliability describes the current observation, but its weighted distances compare that observation against stored appearance. The trend is therefore consistent with more reliable memories making current modality preferences more useful. Under Relative+Abs., Gate also exceeds fixed $\alpha=0.9$ by 0.428 HOTA, 1.146 AssA, and 1.024 IDF1, outperforming the tested fixed-memory settings. Together, the results support complementary roles for absolute calibration, relative weighting, and reliability-guided memory.

\subsection{Reliability Validity}
\label{subsec:reliability_analysis}

Tracking improvements alone do not establish whether reliability reflects useful identity evidence. We therefore conduct the post-hoc analyses in Fig.~\ref{fig:reliability_validity} on VT-MOT with PFTrack detections, without altering the evaluated tracking outputs.

\textbf{Absolute reliability and observation difficulty.}
Using Tri, we collect 548,689 observations from all 120 test sequences and match them to ground truth at IoU $\geq0.5$. Across five equal-width $q_{\mathrm{app}}$ bins, the match rate increases monotonically from 66.84\% to 99.44\%. Low reliability thus concentrates poorly detected or localized observations. Since GT matching also reflects detection quality, this supports $q_{\mathrm{app}}$ as an observation-difficulty cue, not a causal measure of association correctness.

\textbf{Relative reliability and modality discriminability.}
After excluding observations without valid positive memory or a negative competitor, 476,836 samples remain. For modality $b$, define the separation margin $m_b=d_{\mathrm{neg},b}-d_{\mathrm{pos},b}$ using distances to the correct memory and closest incorrect identity. Within each synchronized observation, we compare
\begin{equation}
\Delta q=q_{\mathrm I}-q_{\mathrm R},
\qquad
\Delta m=m_{\mathrm I}-m_{\mathrm R}.
\label{eq:exp_relative_quality_margin}
\end{equation}
The target, frame, and candidates are fixed; agreement between these differences indicates correct modality ordering. Their Spearman correlation is 0.227 overall and 0.278 on the hardest 20\%, selected by the smallest reliability-free uniform tri-branch margins. This selection is independent of reliability, and the positive correlations support informative, though imperfect, modality ordering.

On the hardest 20\%, increasing the mean reliability gap from 0.046 to 0.525 raises the preferred modality's win rate from 53.52\% to 66.58\% and margin advantage from 0.0087 to 0.0420. Stronger gaps therefore provide stronger preference evidence, supporting soft weighting rather than a hard switch.

Preferred-modality Top-1 retrieval also improves under both EMA memory and a bank of the 20 latest features. Here, the modality with higher reliability is compared with the other modality; this diagnostic evaluates modality ordering, while RCA-MOT itself retains soft three-branch weighting. On the hardest 10\%, retrieval increases from 87.4\% to 90.7\% with EMA and from 79.0\% to 84.9\% with the feature bank. Gains are smaller overall and increase on harder subsets, consistent with reliability being most useful when identity evidence is ambiguous.

\subsection{Qualitative Results}
\label{subsec:qualitative_results}

Fig.~\ref{fig:qualitative_results} compares PFTrack with our best-performing RCA-MOT configuration on VT-MOT. Under illumination changes, partial occlusion, and nearby-object ambiguity, panels (a)--(c) show identity switches or inconsistent identity preservation in PFTrack, whereas RCA-MOT maintains the corresponding identities. In panel (d), PFTrack loses the target during the illustrated interval while RCA-MOT maintains and recovers its trajectory identity. These examples illustrate more stable identity preservation in challenging RGB--infrared scenes.
\FloatBarrier
\section{Conclusion}
\label{sec:conclusion}

This work presented RCA-MOT, a reliability-consistent appearance modeling
framework for RGB--infrared multi-object tracking. RCA-MOT retains
RGB-specific, infrared-specific, and fused identity representations, estimates
observation-level branch reliability, and uses the resulting control signal
consistently in both current association and appearance-memory update. Controlled
experiments across two benchmarks and two fixed detection sources showed that the
main improvements are concentrated in identity association. Additional analyses
further indicated that the estimated reliability reflects observation difficulty
and contains modality-specific information related to identity discriminability.
These results support reliability-consistent appearance organization as a simple
and effective complement to increasingly sophisticated multimodal representation
learning.

\balance
\bibliographystyle{IEEEtran}
\bibliography{references}

@article{pftrack,
  author        = {Yabin Zhu and Qianwu Wang and Chenglong Li and Jin Tang and Chengjie Gu and Zhixiang Huang},
  title         = {Visible--thermal multiple object tracking: Large-scale video dataset and progressive fusion approach},
  journal       = {Pattern Recognit.},
  volume        = {161},
  year          = {2025},
  doi           = {10.1016/j.patcog.2024.111330},
  note          = {Art. no. 111330}
}

@article{unirtl,
  author        = {Lian Zhang and Lingxue Wang and Yuzhen Wu and Mingkun Chen and Dezhi Zheng and Liangcai Cao and Bangze Zeng and Yi Cai},
  title         = {{UniRTL}: A universal {RGBT} and low-light benchmark for object tracking},
  journal       = {Pattern Recognit.},
  volume        = {158},
  year          = {2025},
  doi           = {10.1016/j.patcog.2024.110984},
  note          = {Art. no. 110984}
}

@inproceedings{li2019manet,
  author        = {Chenglong Li and Andong Lu and Ai Hua Zheng and Zhengzheng Tu and Jin Tang},
  title         = {Multi-Adapter {RGBT} Tracking},
  booktitle     = {Proc. IEEE/CVF Int. Conf. Comput. Vis. Workshops (ICCVW)},
  pages         = {2262--2270},
  year          = {2019},
  doi           = {10.1109/iccvw.2019.00279}
}

@inproceedings{li2020cat,
  author        = {Chenglong Li and Lei Liu and Andong Lu and Qing Ji and Jin Tang},
  title         = {Challenge-Aware {RGBT} Tracking},
  booktitle     = {Computer Vision -- ECCV},
  pages         = {222--237},
  year          = {2020},
  doi           = {10.1007/978-3-030-58542-6_14}
}

@article{li2022lasher,
  author        = {Chenglong Li and Wanlin Xue and Yaqing Jia and Zhichen Qu and Bin Luo and Jin Tang and Dengdi Sun},
  title         = {{LasHeR}: A Large-Scale High-Diversity Benchmark for {RGBT} Tracking},
  journal       = {IEEE Trans. Image Process.},
  volume        = {31},
  pages         = {392--404},
  year          = {2022},
  doi           = {10.1109/tip.2021.3130533}
}

@article{zhu2021fanet,
  author        = {Yabin Zhu and Chenglong Li and Jin Tang and Bin Luo},
  title         = {Quality-Aware Feature Aggregation Network for Robust {RGBT} Tracking},
  journal       = {IEEE Trans. Intell. Veh.},
  volume        = {6},
  number        = {1},
  pages         = {121--130},
  year          = {2021},
  doi           = {10.1109/tiv.2020.2980735}
}

@inproceedings{zhang2022vtuav,
  author        = {Pengyu Zhang and Jie Zhao and Dong Wang and Huchuan Lu and Xiang Ruan},
  title         = {Visible-Thermal {UAV} Tracking: A Large-Scale Benchmark and New Baseline},
  booktitle     = {Proc. IEEE/CVF Conf. Comput. Vis. Pattern Recognit. (CVPR)},
  pages         = {8876--8885},
  year          = {2022},
  doi           = {10.1109/cvpr52688.2022.00868}
}

@inproceedings{zhu2023vipt,
  author        = {Jiawen Zhu and Simiao Lai and Xin Chen and Dong Wang and Huchuan Lu},
  title         = {Visual Prompt Multi-Modal Tracking},
  booktitle     = {Proc. IEEE/CVF Conf. Comput. Vis. Pattern Recognit. (CVPR)},
  pages         = {9516--9526},
  year          = {2023},
  doi           = {10.1109/cvpr52729.2023.00918}
}

@inproceedings{hui2023tbsi,
  author        = {Tianrui Hui and Zizheng Xun and Fengguang Peng and Junshi Huang and Xiaoming Wei and Xiaolin Wei and Jiao Dai and Jizhong Han and Si Liu},
  title         = {Bridging Search Region Interaction with Template for {RGB-T} Tracking},
  booktitle     = {Proc. IEEE/CVF Conf. Comput. Vis. Pattern Recognit. (CVPR)},
  pages         = {13630--13639},
  year          = {2023},
  doi           = {10.1109/cvpr52729.2023.01310}
}

@inproceedings{qatrgbt,
  author        = {Lei Liu and Chenglong Li and Yun Xiao and Jin Tang},
  title         = {Quality-Aware {RGBT} Tracking via Supervised Reliability Learning and Weighted Residual Guidance},
  booktitle     = {Proc. ACM Int. Conf. Multimedia (ACM MM)},
  pages         = {3129--3137},
  year          = {2023},
  doi           = {10.1145/3581783.3612341}
}

@inproceedings{hou2024sdstrack,
  author        = {Xiaojun Hou and Jiazheng Xing and Yijie Qian and Yaowei Guo and Shuo Xin and Junhao Chen and Kai Tang and Mengmeng Wang and Zhengkai Jiang and Liang Liu and Yong Liu},
  title         = {{SDSTrack}: Self-Distillation Symmetric Adapter Learning for Multi-Modal Visual Object Tracking},
  booktitle     = {Proc. IEEE/CVF Conf. Comput. Vis. Pattern Recognit. (CVPR)},
  pages         = {26541--26551},
  year          = {2024},
  doi           = {10.1109/cvpr52733.2024.02507}
}

@inproceedings{xiao2025caformer,
  author        = {Yun Xiao and Jiacong Zhao and Andong Lu and Chenglong Li and Bing Yin and Yin Lin and Cong Liu},
  title         = {Cross-modulated Attention Transformer for {RGBT} Tracking},
  booktitle     = {Proc. AAAI Conf. Artif. Intell.},
  volume        = {39},
  number        = {8},
  pages         = {8682--8690},
  year          = {2025},
  doi           = {10.1609/aaai.v39i8.32938}
}

@article{lu2025dmcnet,
  author        = {Andong Lu and Cun Qian and Chenglong Li and Jin Tang and Liang Wang},
  title         = {Duality-Gated Mutual Condition Network for {RGBT} Tracking},
  journal       = {IEEE Trans. Neural Netw. Learn. Syst.},
  volume        = {36},
  number        = {3},
  pages         = {4118--4131},
  year          = {2025},
  doi           = {10.1109/tnnls.2022.3157594}
}

@inproceedings{bewley2016sort,
  author        = {Alex Bewley and Zongyuan Ge and Lionel Ott and Fabio Ramos and Ben Upcroft},
  title         = {Simple online and realtime tracking},
  booktitle     = {Proc. IEEE Int. Conf. Image Process. (ICIP)},
  pages         = {3464--3468},
  year          = {2016},
  doi           = {10.1109/icip.2016.7533003}
}

@inproceedings{deepsort,
  author        = {Nicolai Wojke and Alex Bewley and Dietrich Paulus},
  title         = {Simple online and realtime tracking with a deep association metric},
  booktitle     = {Proc. IEEE Int. Conf. Image Process. (ICIP)},
  pages         = {3645--3649},
  year          = {2017},
  doi           = {10.1109/icip.2017.8296962}
}

@article{zhang2021fairmot,
  author        = {Yifu Zhang and Chunyu Wang and Xinggang Wang and Wenjun Zeng and Wenyu Liu},
  title         = {{FairMOT}: On the Fairness of Detection and Re-identification in Multiple Object Tracking},
  journal       = {Int. J. Comput. Vis.},
  volume        = {129},
  number        = {11},
  pages         = {3069--3087},
  year          = {2021},
  doi           = {10.1007/s11263-021-01513-4}
}

@inproceedings{zhou2020centertrack,
  author        = {Xingyi Zhou and Vladlen Koltun and Philipp Kr{\"a}henb{\"u}hl},
  title         = {Tracking Objects as Points},
  booktitle     = {Computer Vision -- ECCV},
  pages         = {474--490},
  year          = {2020},
  doi           = {10.1007/978-3-030-58548-8_28}
}

@inproceedings{wu2021trades,
  author        = {Jialian Wu and Jiale Cao and Liangchen Song and Yu Wang and Ming Yang and Junsong Yuan},
  title         = {Track to Detect and Segment: An Online Multi-Object Tracker},
  booktitle     = {Proc. IEEE/CVF Conf. Comput. Vis. Pattern Recognit. (CVPR)},
  pages         = {12347--12356},
  year          = {2021},
  doi           = {10.1109/cvpr46437.2021.01217}
}

@article{sun2020transtrack,
  author        = {Peize Sun and Jinkun Cao and Yi Jiang and Rufeng Zhang and Enze Xie and Zehuan Yuan and Changhu Wang and Ping Luo},
  title         = {{TransTrack}: Multiple Object Tracking with Transformer},
  journal       = {arXiv preprint arXiv:2012.15460},
  year          = {2020},
  doi           = {10.48550/arxiv.2012.15460},
  eprint        = {2012.15460},
  archivePrefix = {arXiv}
}

@inproceedings{bytetrack,
  author        = {Yifu Zhang and Peize Sun and Yi Jiang and Dongdong Yu and Fucheng Weng and Zehuan Yuan and Ping Luo and Wenyu Liu and Xinggang Wang},
  title         = {{ByteTrack}: Multi-object Tracking by Associating Every Detection Box},
  booktitle     = {Computer Vision -- ECCV},
  pages         = {1--21},
  year          = {2022},
  doi           = {10.1007/978-3-031-20047-2_1}
}

@inproceedings{ocsort,
  author        = {Jinkun Cao and Jiangmiao Pang and Xinshuo Weng and Rawal Khirodkar and Kris Kitani},
  title         = {Observation-Centric {SORT}: Rethinking {SORT} for Robust Multi-Object Tracking},
  booktitle     = {Proc. IEEE/CVF Conf. Comput. Vis. Pattern Recognit. (CVPR)},
  pages         = {9686--9696},
  year          = {2023},
  doi           = {10.1109/cvpr52729.2023.00934}
}

@article{du2022strongsort,
  author        = {Yunhao Du and Zhicheng Zhao and Yang Song and Yanyun Zhao and Fei Su and Tao Gong and Hongying Meng},
  title         = {{StrongSORT}: Make {DeepSORT} Great Again},
  journal       = {IEEE Trans. Multimedia},
  volume        = {25},
  pages         = {8725--8737},
  year          = {2023},
  doi           = {10.1109/tmm.2023.3240881}
}

@article{aharon2022botsort,
  author        = {Nir Aharon and Roy Orfaig and Ben-Zion Bobrovsky},
  title         = {{BoT-SORT}: Robust Associations Multi-Pedestrian Tracking},
  journal       = {arXiv preprint arXiv:2206.14651},
  year          = {2022},
  doi           = {10.48550/arxiv.2206.14651},
  eprint        = {2206.14651},
  archivePrefix = {arXiv}
}

@inproceedings{maggiolino2023deepocsort,
  author        = {Gerard Maggiolino and Adnan Ahmad and Jinkun Cao and Kris Kitani},
  title         = {{Deep OC-Sort}: Multi-Pedestrian Tracking by Adaptive Re-Identification},
  booktitle     = {Proc. IEEE Int. Conf. Image Process. (ICIP)},
  pages         = {3025--3029},
  year          = {2023},
  doi           = {10.1109/icip49359.2023.10222576}
}

@inproceedings{hybridsort,
  author        = {Mingzhan Yang and Guangxin Han and Bin Yan and Wenhua Zhang and Jinqing Qi and Huchuan Lu and Dong Wang},
  title         = {{Hybrid-SORT}: Weak Cues Matter for Online Multi-Object Tracking},
  booktitle     = {Proc. AAAI Conf. Artif. Intell.},
  volume        = {38},
  number        = {7},
  pages         = {6504--6512},
  year          = {2024},
  doi           = {10.1609/aaai.v38i7.28471}
}

@inproceedings{cui2023sportsmot,
  author        = {Yutao Cui and Chenkai Zeng and Xiaoyu Zhao and Yichun Yang and Gangshan Wu and Limin Wang},
  title         = {{SportsMOT}: A Large Multi-Object Tracking Dataset in Multiple Sports Scenes},
  booktitle     = {Proc. IEEE/CVF Int. Conf. Comput. Vis. (ICCV)},
  pages         = {9887--9897},
  year          = {2023},
  doi           = {10.1109/iccv51070.2023.00910}
}

@article{lv2024pidmot,
  author        = {Weiyi Lv and Ning Zhang and Junjie Zhang and Dan Zeng},
  title         = {One-Shot Multiple Object Tracking With Robust {ID} Preservation},
  journal       = {IEEE Trans. Circuits Syst. Video Technol.},
  volume        = {34},
  number        = {6},
  pages         = {4473--4488},
  year          = {2024},
  doi           = {10.1109/tcsvt.2023.3339609}
}

@inproceedings{meinhardt2022trackformer,
  author        = {Tim Meinhardt and Alexander Kirillov and Laura Leal-Taix{\'e} and Christoph Feichtenhofer},
  title         = {{TrackFormer}: Multi-Object Tracking with Transformers},
  booktitle     = {Proc. IEEE/CVF Conf. Comput. Vis. Pattern Recognit. (CVPR)},
  pages         = {8834--8844},
  year          = {2022},
  doi           = {10.1109/cvpr52688.2022.00864}
}

@inproceedings{zeng2022motr,
  author        = {Fangao Zeng and Bin Dong and Yuang Zhang and Tiancai Wang and Xiangyu Zhang and Yichen Wei},
  title         = {{MOTR}: End-to-End Multiple-Object Tracking with Transformer},
  booktitle     = {Computer Vision -- ECCV},
  pages         = {659--675},
  year          = {2022},
  doi           = {10.1007/978-3-031-19812-0_38}
}

@inproceedings{zhang2023motrv2,
  author        = {Yuang Zhang and Tiancai Wang and Xiangyu Zhang},
  title         = {{MOTRv2}: Bootstrapping End-to-End Multi-Object Tracking by Pretrained Object Detectors},
  booktitle     = {Proc. IEEE/CVF Conf. Comput. Vis. Pattern Recognit. (CVPR)},
  pages         = {22056--22065},
  year          = {2023},
  doi           = {10.1109/cvpr52729.2023.02112}
}

@inproceedings{pang2021qdtrack,
  author        = {Jiangmiao Pang and Linlu Qiu and Xia Li and Haofeng Chen and Qi Li and Trevor Darrell and Fisher Yu},
  title         = {Quasi-Dense Similarity Learning for Multiple Object Tracking},
  booktitle     = {Proc. IEEE/CVF Conf. Comput. Vis. Pattern Recognit. (CVPR)},
  pages         = {164--173},
  year          = {2021},
  doi           = {10.1109/cvpr46437.2021.00023}
}

@inproceedings{seidenschwarz2023ghost,
  author        = {Jenny Seidenschwarz and Guillem Bras{\'o} and Victor Castro Serrano and Ismail Elezi and Laura Leal-Taix{\'e}},
  title         = {Simple Cues Lead to a Strong Multi-Object Tracker},
  booktitle     = {Proc. IEEE/CVF Conf. Comput. Vis. Pattern Recognit. (CVPR)},
  pages         = {13813--13823},
  year          = {2023},
  doi           = {10.1109/cvpr52729.2023.01327}
}

@inproceedings{gao2023memotr,
  author        = {Ruopeng Gao and Limin Wang},
  title         = {{MeMOTR}: Long-Term Memory-Augmented Transformer for Multi-Object Tracking},
  booktitle     = {Proc. IEEE/CVF Int. Conf. Comput. Vis. (ICCV)},
  pages         = {9867--9876},
  year          = {2023},
  doi           = {10.1109/iccv51070.2023.00908}
}

@inproceedings{luo2019bagtricks,
  author        = {Hao Luo and Youzhi Gu and Xingyu Liao and Shenqi Lai and Wei Jiang},
  title         = {Bag of Tricks and a Strong Baseline for Deep Person Re-Identification},
  booktitle     = {Proc. IEEE/CVF Conf. Comput. Vis. Pattern Recognit. Workshops (CVPRW)},
  pages         = {1487--1495},
  year          = {2019},
  doi           = {10.1109/cvprw.2019.00190}
}

@article{hermans2017triplet,
  author        = {Alexander Hermans and Lucas Beyer and Bastian Leibe},
  title         = {In Defense of the Triplet Loss for Person Re-Identification},
  journal       = {arXiv preprint arXiv:1703.07737},
  year          = {2017},
  doi           = {10.48550/arxiv.1703.07737},
  eprint        = {1703.07737},
  archivePrefix = {arXiv}
}

@inproceedings{sun2020circleloss,
  author        = {Yifan Sun and Changmao Cheng and Yuhan Zhang and Chi Zhang and Liang Zheng and Zhongdao Wang and Yichen Wei},
  title         = {Circle Loss: A Unified Perspective of Pair Similarity Optimization},
  booktitle     = {Proc. IEEE/CVF Conf. Comput. Vis. Pattern Recognit. (CVPR)},
  pages         = {6397--6406},
  year          = {2020},
  doi           = {10.1109/cvpr42600.2020.00643}
}

@inproceedings{liu2017qan,
  author        = {Yu Liu and Junjie Yan and Wanli Ouyang},
  title         = {Quality Aware Network for Set to Set Recognition},
  booktitle     = {Proc. IEEE Conf. Comput. Vis. Pattern Recognit. (CVPR)},
  pages         = {4694--4703},
  year          = {2017},
  doi           = {10.1109/cvpr.2017.499}
}

@inproceedings{he2021transreid,
  author        = {Shuting He and Hao Luo and Pichao Wang and Fan Wang and Hao Li and Wei Jiang},
  title         = {{TransReID}: Transformer-based Object Re-Identification},
  booktitle     = {Proc. IEEE/CVF Int. Conf. Comput. Vis. (ICCV)},
  pages         = {14993--15002},
  year          = {2021},
  doi           = {10.1109/iccv48922.2021.01474}
}

@article{bernardin2008clearmot,
  author        = {Keni Bernardin and Rainer Stiefelhagen},
  title         = {Evaluating Multiple Object Tracking Performance: The {CLEAR MOT} Metrics},
  journal       = {EURASIP J. Image Video Process.},
  volume        = {2008},
  pages         = {1--10},
  year          = {2008},
  doi           = {10.1155/2008/246309}
}

@article{mctrack,
  author        = {Jianbo Ma and Hui Luo and Shuaicheng Niu and Peilin Zhao and Yunfeng Liu and Yuxing Wei and Jianlin Zhang},
  title         = {Multi-Stage Cross-Modality Feature Interaction for {RGB}-Thermal Multi-Object Tracking},
  journal       = {IEEE Trans. Circuits Syst. Video Technol.},
  volume        = {36},
  number        = {2},
  pages         = {2449--2463},
  year          = {2026},
  doi           = {10.1109/TCSVT.2025.3612499}
}

@inproceedings{ristani2016idmetrics,
  author        = {Ergys Ristani and Francesco Solera and Roger Zou and Rita Cucchiara and Carlo Tomasi},
  title         = {Performance Measures and a Data Set for Multi-target, Multi-camera Tracking},
  booktitle     = {Computer Vision -- ECCV Workshops},
  pages         = {17--35},
  year          = {2016},
  doi           = {10.1007/978-3-319-48881-3_2}
}

@article{luiten2021hota,
  author        = {Jonathon Luiten and Aljo{\v{s}}a O{\v{s}}ep and Patrick Dendorfer and Philip Torr and Andreas Geiger and Laura Leal-Taix{\'e} and Bastian Leibe},
  title         = {{HOTA}: A Higher Order Metric for Evaluating Multi-object Tracking},
  journal       = {Int. J. Comput. Vis.},
  volume        = {129},
  number        = {2},
  pages         = {548--578},
  year          = {2021},
  doi           = {10.1007/s11263-020-01375-2}
}

@article{dendorfer2021motchallenge,
  author        = {Patrick Dendorfer and Aljo{\v{s}}a O{\v{s}}ep and Anton Milan and Konrad Schindler and Daniel Cremers and Ian Reid and Stefan Roth and Laura Leal-Taix{\'e}},
  title         = {{MOTChallenge}: A Benchmark for Single-Camera Multiple Target Tracking},
  journal       = {Int. J. Comput. Vis.},
  volume        = {129},
  number        = {4},
  pages         = {845--881},
  year          = {2021},
  doi           = {10.1007/s11263-020-01393-0}
}

@inproceedings{wang2020multimodalhard,
  author        = {Weiyao Wang and Du Tran and Matt Feiszli},
  title         = {What Makes Training Multi-Modal Classification Networks Hard?},
  booktitle     = {Proc. IEEE/CVF Conf. Comput. Vis. Pattern Recognit. (CVPR)},
  pages         = {12692--12702},
  year          = {2020},
  doi           = {10.1109/cvpr42600.2020.01271}
}

@inproceedings{peng2022ogmge,
  author        = {Xiaokang Peng and Yake Wei and Andong Deng and Dong Wang and Di Hu},
  title         = {Balanced Multimodal Learning via On-the-fly Gradient Modulation},
  booktitle     = {Proc. IEEE/CVF Conf. Comput. Vis. Pattern Recognit. (CVPR)},
  pages         = {8228--8237},
  year          = {2022},
  doi           = {10.1109/cvpr52688.2022.00806}
}

@article{neverova2016moddrop,
  author        = {Natalia Neverova and Christian Wolf and Graham Taylor and Florian Nebout},
  title         = {{ModDrop}: Adaptive Multi-Modal Gesture Recognition},
  journal       = {IEEE Trans. Pattern Anal. Mach. Intell.},
  volume        = {38},
  number        = {8},
  pages         = {1692--1706},
  year          = {2016},
  doi           = {10.1109/tpami.2015.2461544}
}

@article{hinton2015distilling,
  author        = {Geoffrey Hinton and Oriol Vinyals and Jeff Dean},
  title         = {Distilling the Knowledge in a Neural Network},
  journal       = {arXiv preprint arXiv:1503.02531},
  year          = {2015},
  doi           = {10.48550/arxiv.1503.02531},
  eprint        = {1503.02531},
  archivePrefix = {arXiv}
}

\end{document}